\documentclass[Journal,NoLineNumbers,InsideFigs]{ascelike-new}
\usepackage[utf8]{inputenc}
\usepackage[T1]{fontenc}
\usepackage{lmodern}
\usepackage{graphicx}
\usepackage[style=base,figurename=Fig.,labelfont=bf,labelsep=period]{caption}
\usepackage{subcaption}
\usepackage{amsmath}
\usepackage{booktabs}
\usepackage{siunitx} 
\usepackage{hologo}
\usepackage[utf8]{inputenc}
\usepackage{float}
\usepackage{algorithm}
\usepackage{algpseudocode}
\usepackage{booktabs}
\usepackage{tabularx}
\usepackage{comment}
\usepackage{csquotes} 
\usepackage{xcolor}
\usepackage{amsmath,amssymb,amsfonts}
\usepackage{graphicx}
\usepackage{textcomp}
\usepackage{adjustbox}
\usepackage{placeins}
\usepackage{newtxtext,newtxmath}
\newcommand{\OPEN}{Open}
\usepackage[colorlinks=true,citecolor=red,linkcolor=black]{hyperref}
\NameTag{Amani, \today}
\begin{document}

\title{Digital Twin‑Guided Robot Path Planning: A Beta–Bernoulli Fusion with Large Language Model as a Sensor}

\author[1,2]{Mani Amani}
\author[1]{Reza Akhavian}

\affil[1]{Department of Civil, Construction, and Environmental Engineering, San Diego State University, San Diego, CA, United States}
\affil[2]{Department of Electrical and Computer Engineering, University of California, San Diego, San Diego, CA, United States, with corresponding author email. Email: rakhavian@sdsu.edu}

\maketitle
\begin{abstract}
Integrating natural language (NL) prompts into robotic mission planning has attracted significant interest in recent years. In the construction domain, Building Information Models (BIM) encapsulate rich NL descriptions of the environment. We present a novel framework that fuses NL directives with BIM–derived semantic maps via a Beta–Bernoulli Bayesian fusion by interpreting the LLM as a sensor: each obstacle’s design‐time repulsive coefficient is treated as a \(\mathrm{Beta}(\alpha,\beta)\) random variable and LLM‐returned “danger scores” are incorporated as pseudo‐counts to update \(\alpha\) and \(\beta\).  The resulting posterior mean yields a continuous, context‐aware repulsive gain that augments a Euclidean‐distance‐based potential field for cost heuristics. By adjusting gains based on sentiment and context inferred from user prompts, our method guides robots along safer, more context‐aware paths. This provides a numerically stable method that can chain multiple natural commands and prompts from construction workers and foreman to enable planning while giving flexibility to be integrated in any learned or classical AI framework. Simulation results demonstrate that this Beta–Bernoulli fusion yields both qualitative and quantitative improvements in path robustness and validity. Our code is publicly available on GitHub.
\end{abstract}
\begin{figure}[t]
    \centering
    \includegraphics[width=\columnwidth]{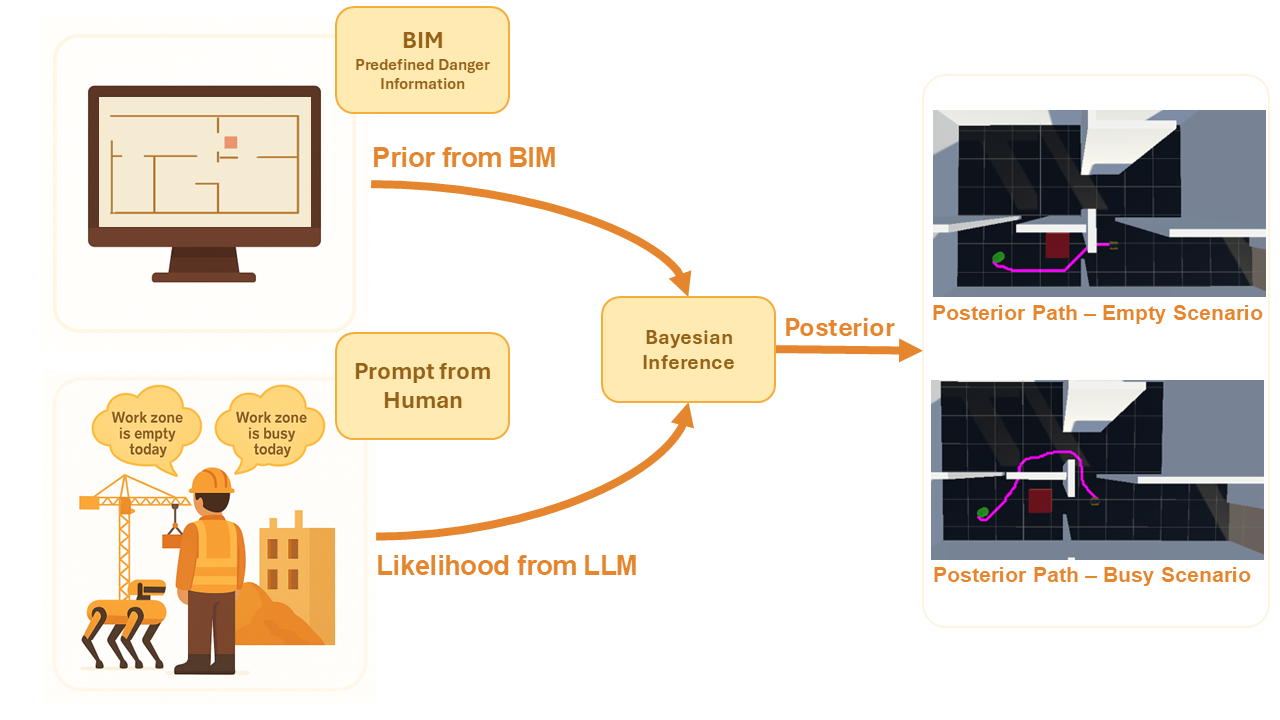}
    \caption{Overview of the proposed BIM–LLM Bayesian fusion framework for danger-aware robotic mission planning.}
    \label{fig:workflow}
\end{figure}

\section{Introduction}
The emergence of large language models (LLMs) has fundamentally transformed natural language processing (NLP) capabilities across diverse domains, with robotic planning experiencing similar paradigm shifts through enhanced semantic understanding and contextual reasoning \cite{UCSDLLM}\cite{LLMsgrowing}. While there has been a concentrated focus on the advancement of prompted robotic applications, most of the research on language-commanded robotics is limited to predefined objects and primitives \cite{primitives}. This restriction confines robots to a limited set of predefined actions. Sometimes, mission parameters must be adjusted and tuned given unseen and undefined prompts. Furthermore, it is well established that humans expect their collaborators, including AI, to interpret implicit cues beyond explicit wording \cite{humanAI}. This creates a need for a more flexible framework to analyze implicit sentiment information from natural language prompts. Our proposed BIM–LLM Bayesian fusion framework, illustrated in Figure \ref{fig:workflow}, addresses this gap.
Since implicit and semantic information inherently depend on subjective human interpretation, a probabilistic framework provides a principled approach to quantify and propagate these uncertainties through the planning system.  Specifically, the inherent ambiguity and context-dependency of natural language sentiment, combined with the possibility of LLM hallucinations, require a stable probabilistic approach. Therefore, leveraging deep learning models is considered a suitable approach since they have been shown to have high-accuracy sentiment analysis \cite{sentimentanlasysiukrain}. LLMs have demonstrated extensive success in sentiment analysis, positioning them as attractive candidates for assessing Theory of Mind-like associations \cite{LLMSenitmentanalysis}. Theory of Mind refers to the ability to attribute mental states (like beliefs, desires, and intentions) to oneself and others. In this context, the 'Theory of Mind-like associations' means that the system is designed to infer and relate implicit emotional or cognitive states from verbal communication and, more generally, observed behavior \cite{ToM}. While these advancements show promise in human-machine interaction, in robotics, the analysis must operate under strict mission constraints and heuristics, requiring that the sentiment data be transformed into actionable insights that can be adapted across various scenarios. This introduces additional complexity in transforming raw sentiment data into an actionable and generalizable format across diverse mission scenarios. Furthermore, for practical applications, it is essential to establish a general mapping between sentiment cues and environmental context to ensure accurate robot operations.

\par
Semantic maps have been extensively studied as a means to integrate qualitative and spatial information \cite{SemanticMap}. Typically, these maps assign a class label and descriptive textual information to spatial objects, enabling real-time map generation and planning. In many applied settings, there are existing data formats analogous to semantic maps. For example, BIM is a widely used format that represents both the three-dimensional geometry and associated textual information of a built environment \cite{BIN}. Given its dual representation of spatial and semantic data, BIM serves as an excellent analog to a semantic map. Indeed, previous research has successfully leveraged BIM to provide the semantic information for robot planning \cite{KAMATLLM}. Furthermore, the integration of internet of things (IoT) sensors, real-time data streams, and machine learning has driven an evolution from static BIM representations toward dynamic Digital Twins that continuously update to reflect current site conditions, enabling more responsive robotic navigation in changing construction environments. 
\par 

Previous works typically treat the LLM as the primary planning module, leveraging environmental information to generate semantically meaningful plans \cite{LLMplanner}. 
In contrast, we propose an alternative interpretation: using the LLM as a sensor rather than a planner. In this framework, the LLM provides semantic information that can be embedded into the map or any intermediate representation used during planning. The planning module itself remains agnostic to the LLM, allowing for the use of any classical or learning-based planner in conjunction with the LLM-as-sensor approach. Although visual–language maps have shown great promise \cite{NotSemantic}, most prior work focuses on robot planning from direct commands. In contrast, our contribution demonstrates that LLMs, combined with probabilistic map updates, can account for contextual and implicit sentiments that are not explicitly expressed in the natural-language command. 
\par

Using NL as a cost signal for obstacle avoidance has been explored previously \cite{LanguageAScost}. Prior work shows that VLMs can identify dynamic obstacles and assign danger scores. We extend this line of work by incorporating implicit and contextual information about the environment and by enabling the chaining and consolidation of multiple commands, yielding more flexible and general language-guided costmaps. Semantic planning has also been studied in the realm of visual-language-action models (VLA). However, these models are often expensive to train and deploy. Furthermore, these models often do not have formal verification and guarantees regarding their performance \cite{VLAhard}. By incorporating NL data as sensor information that can augment costs, we can integrate NL with classical and certifiable AI algorithms such as A* for path-finding, which is commonly used in robotics \cite{Halder} \cite{UCSDLLM}. In the presented framework, the ideas proposed in studies such as \cite{UCSDLLM} are extended to risk shaping rather than task satisfaction, with BIM‑grounded semantics.
Our contributions in this paper can be summarized as:
\begin{enumerate}
  \item We propose a novel interpretation of LLMs as semantic sensors for robotic navigation, where natural language prompts about construction site conditions update obstacle danger coefficients through Beta-Bernoulli fusion. This approach leverages BIM family semantics to ground language understanding while maintaining map structure invariance.
  \item We derive a closed-form Beta–Bernoulli fusion of prompt-derived danger scores into continuous repulsive gains, thereby formulating object avoidance as a repulsive potential-field cost metric.
  \item We demonstrate that off-the-shelf LLMs correctly interpret domain-specific construction prompts (no domain fine-tuning needed), and that our fusion preserves the bounded-suboptimal guarantees of Multi-Heuristic A*.
\end{enumerate}
\section{Related Works}
Robotic policy learning aided by LLMs has been previously explored. In one study \cite{GAIL}, the authors use Generative Adversarial Imitation Learning (GAIL) to enable the agent to learn reusable skills with a single policy and solve zero-shot tasks under the guidance of LLMs. The natural language task is broken down into actionable commands, which are then used to learn a control policy.
Furthermore, there has been recent research on using LLM-guided robots for safe navigation \cite{INsturctional}. In this work, the authors use LLMs to mission plan using instructions in unknown environments. The authors propose a methodology that parses instructions and semantic constraints (i.e., "avoid repair area") to generate a set of actions to reach the initial high-level instruction. Instruction-based planning has been further extended into multi-agent planning \cite{MA_plan}. In this work, the high-level instruction is decomposed into sub-tasks, from which coalitions are formed. This enables multi-robot instruction using one prompt.
Other studies have explored extending LLMs from conducting step-wise scene and task understanding into partially observable task planning \cite{Partial}. This approach advances the effectiveness of robots in physical open-vocabulary tasks. However, throughout most of these works, implicit sentiment analysis and its effects on mission constraints and behavior have not been explored or addressed.
\par
VLAs such as RT-2 \cite{RT2} work by using natural language and visual information to generate actions that are actionable by the robot. However, these models are notoriously data intensive and require large-scale pretraining on robotic datasets \cite{expensivePretrain}. This becomes a significant issue when dealing with domain-specific tasks \cite{DomainSpecific}. In the case of construction robotics. a safety-critical domain, poor generalization and hallucinations from a VLA model cannot be afforded in practical applications. Moreover, VLAs are often computationally expensive, with billions of parameters in their architectures \cite{paramCount}, creating increasing barriers for investment and development in real-time construction robotics applications. To formulate an effective alternative, our method proposes a closed-form and computationally efficient consolidation of natural language and mission planning, without the need for extensive data training and heavy computation on the edge.
\section{Methodology}
\subsection{EDF}
\begin{figure}[!htbp]
    \centering
    \begin{subfigure}[b]{0.49\columnwidth}
        \centering
        \includegraphics[width=\columnwidth]{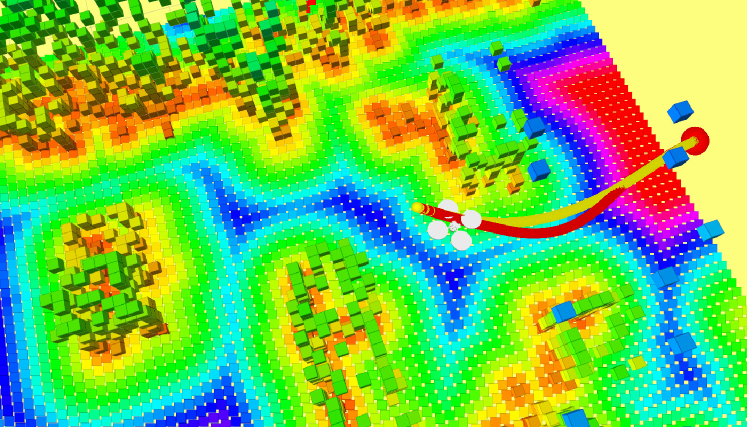}
        \caption{Euclidean distance field from the FIESTA algorithm \cite{fiesta}}
        \label{fig:edf_fiesta}
    \end{subfigure}
    \hfill
    \begin{subfigure}[b]{0.49\columnwidth}
        \centering
        \includegraphics[width=\columnwidth]{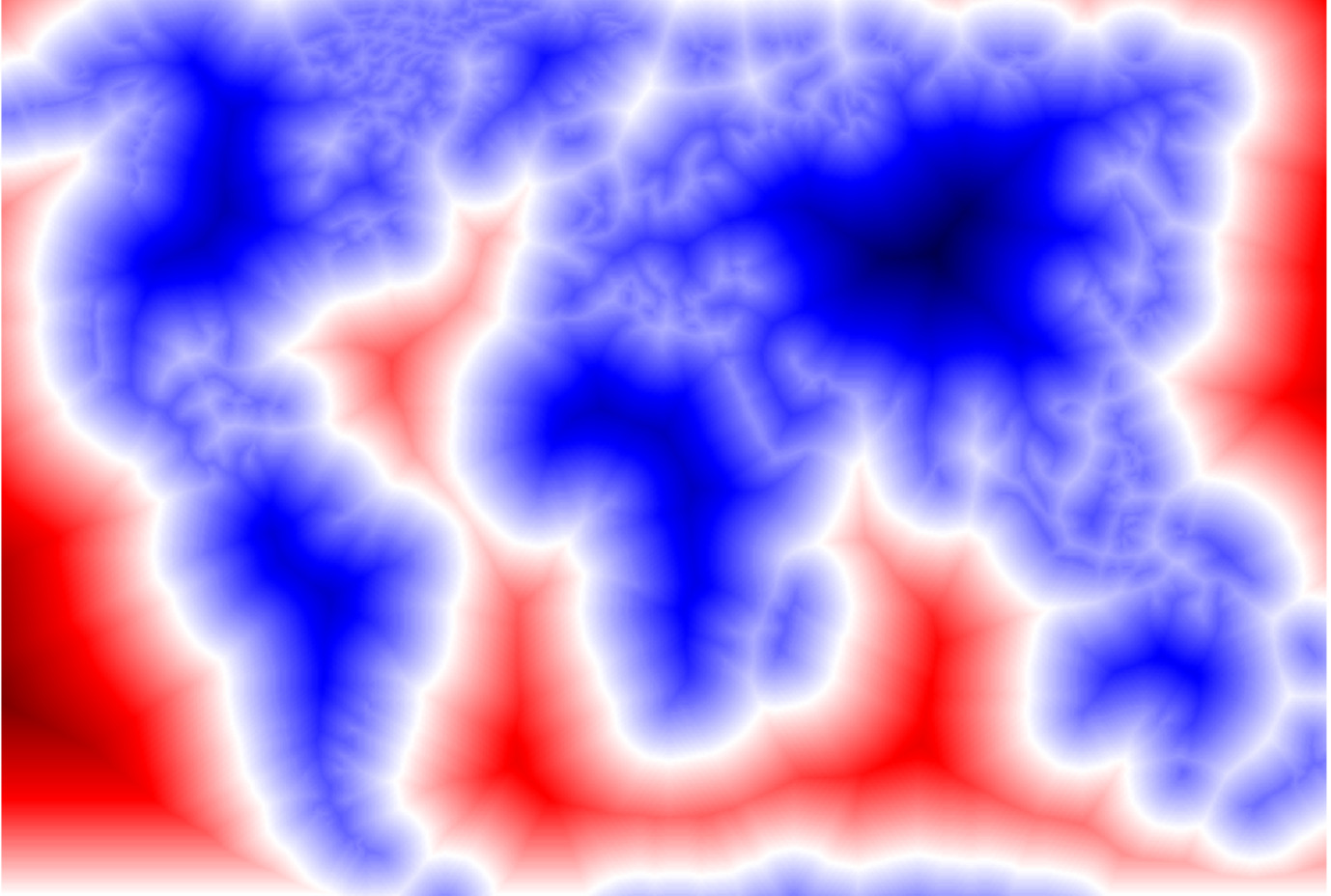}
        \caption{Global world map represented as an EDF of land masses}
        \label{fig:edf_worldmap}
    \end{subfigure}
    \caption{Examples of Euclidean distance fields: (a) output of the FIESTA algorithm, (b) global land‐mass distance field.}
    \label{fig:edf_examples}
\end{figure}

Given that BIMs can be used as a semantic map, we have the benefit of having both spatial and textual information regarding the environment. While there are many types of mapping strategies, distance function mapping is one of the popular continuous representations of the environment. Euclidean distance functions (EDF), signed distance functions (SDF), and signed-directional distance functions (SDDF) are commonly used to enable the representation of the map. Formally,  a distance field is a function \(	\mathbb{R}^d \to\mathbb{R}\) where the position \(\mathbf{p} \in \mathbb{R}^d\) is mapped to a scalar \(d\) is the \(\inf(f(\mathbf{p}))\). For example, an EDF can be defined as 
\begin{equation}
    D_A(\mathbf{x}) = \inf_{\mathbf{y} \in A} \| \mathbf{x} - \mathbf{y} \|_2
\end{equation}
An example of the EDF generation of a map is shown in Figure \ref{fig:edf_examples}.
EDFs are used extensively in robotic applications as representations of free space and obstacles. A larger number would represent free space, and a 0 value would represent an obstacle. Given the spatial positions available in the BIM, we can also generate an EDF to represent a continuous space for robotic mapping. Here, we can use the 2D map of a BIM model and represent it as an EDF.

\par
Since the infimum operator is used in most formulations of distance fields, only the obstacle with the minimum distance is considered in the generation of the distance field map. This can be interpreted as a downside in cluttered environments or areas with poor mapping and obstacle localizations. In construction sites, BIM can be used as an effective analog for the robot map. However, BIMs can have localization errors for design elements, especially when the model is not updated frequently based on the latest design changes. Recent studies have achieved a localization error of 20 cm \cite{BIMLocalationzerpoint2}, which is not considered safe around humans and in job site environments. Depending on the implementation, this error can be detrimental to robotic operations, given the potential for collisions. While there are many different strategies to resolve this problem, we propose using information contained in the NL context to address the avoidance of high-risk areas, such as those that are dynamic, cluttered, or would cause significant disruption to construction operations in the event of a collision. 
\par

We propose to use the concepts from an EDF to generate a new formulation of a distance field, which is a summation of a function of distances, dubbed as repulsive fields, on nodes in space. This would result in larger numbers representing areas of high risk and smaller numbers representing safer areas. Here, we would identify obstacles and their implied safety from the BIM family models and names, and use both the textual data and distance information as costs representing the potential field. Previous works have shown that generating potential fields from BIMs can enhance object avoidance.
\begin{figure}[ht]
  \centering
  \begin{subfigure}[t]{0.48\columnwidth}
    \centering
    \includegraphics[width=\columnwidth]{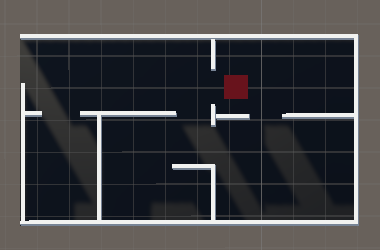}
    \caption{Original BIM environment}
    \label{fig:basebim}
  \end{subfigure}
  \hfill
  \begin{subfigure}[t]{0.48\columnwidth}
    \centering
    \includegraphics[width=\columnwidth]{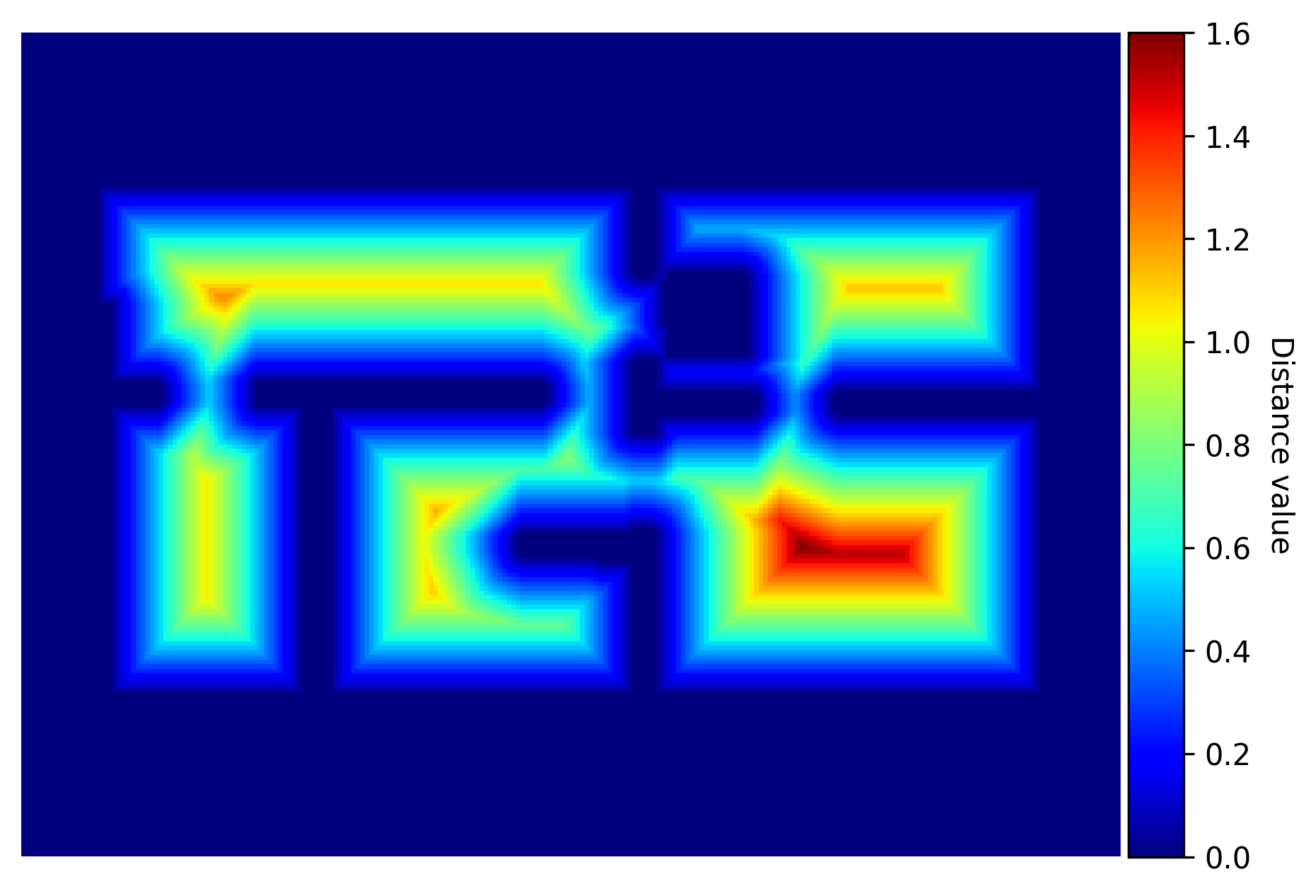}
    \caption{Euclidean distance field (EDF)}
    \label{fig:edf}
  \end{subfigure}
  \caption{Visualization of the BIM environment and its corresponding EDF.}
  \label{fig:bim_edf}
\end{figure}

To improve upon previous works, we plan to use the NL task also in conjunction with the BIM to further augment the task carried out by the robot.

\subsection{Potential Field Formulation}

Given a precomputed distance field and the obstacle geometries provided by BIM families, we define the repulsive potential at any node \((x,y)\) as the inverse exponential of its distance to obstacles:

\begin{equation}
F_{\mathrm{rep},i}(x,y)
= \lambda_i \,\exp\!\bigl(-D_i(x,y)\bigr).
\label{eq:F_rep}
\end{equation}

where \(D_{\mathrm{i}}(x,y)\) is the Euclidean distance to the nearest obstacle and \(\lambda_i\) is the potential gain to tune the obstacle avoidance cost:

\begin{equation}
D_i(x,y)
= \| \mathbf{x} - \mathbf{o}_i \|_2
\label{eq:D_euclid}
\end{equation}

Finally, the total potential across all \(n\) obstacles is the weighted sum of each repulsive term:

\begin{equation}
F_{\mathrm{total}}(x,y)
= \sum_{i=1}^{n} F_{\mathrm{rep},i}(x,y)
= \sum_{i=1}^{n} \lambda_i \, \exp\!\big(-D_i(x,y)\big).
\label{eq:F_total}
\end{equation}

\begin{figure}
    \centering
    \includegraphics[width=1\columnwidth]{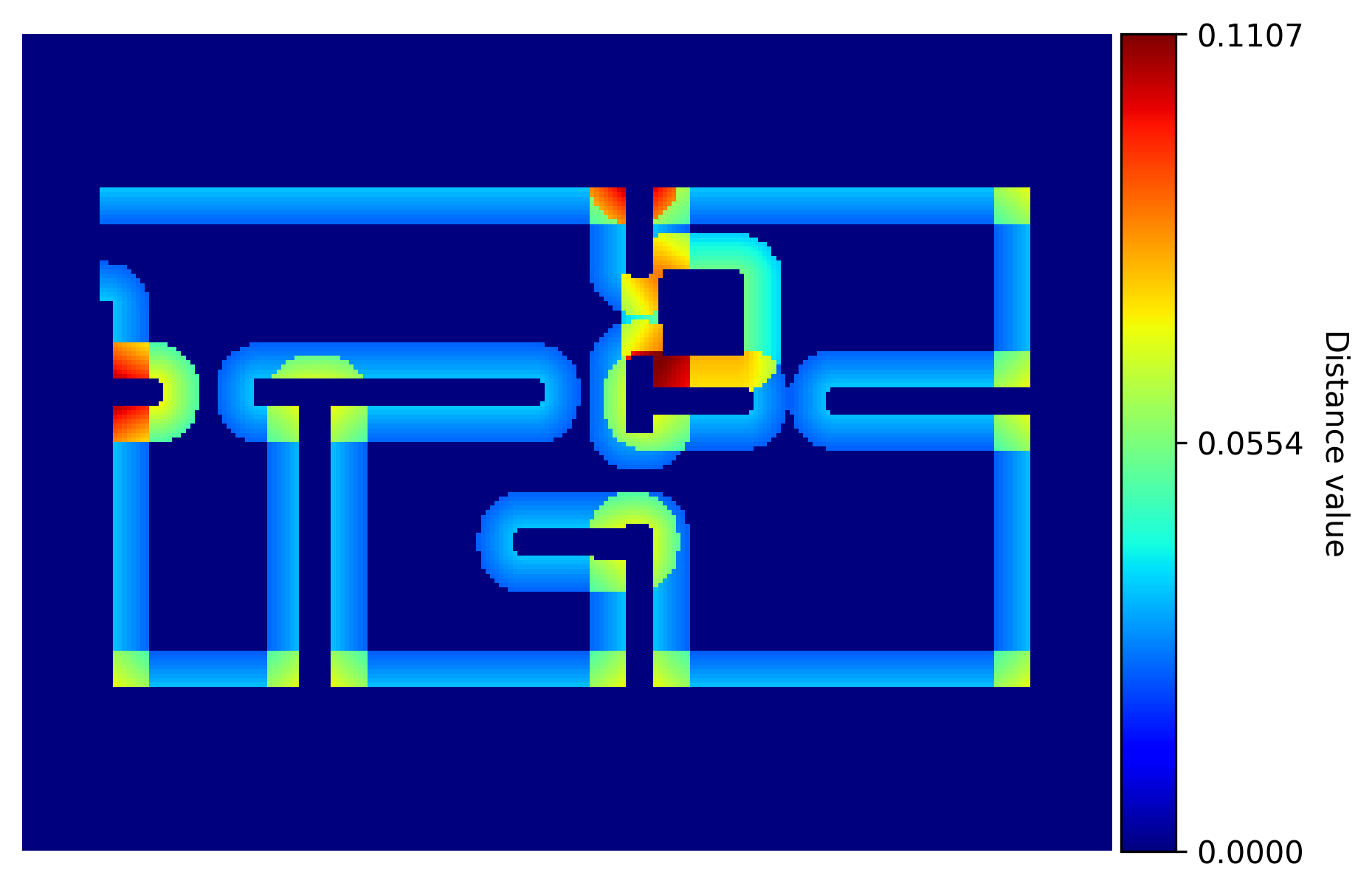}
    \caption{An example of the potential field generated from Equation \ref{eq:F_total}}
    \label{fig:ExPot}
\end{figure}

An example of the resulting potential field can be seen in Figure \ref{fig:ExPot}.

\subsection{Multi-Heuristic A*}
\subsubsection{Dynamic-Programming Formulation}
We can view the joint optimization of the distance and potential field cost as a classic dynamic‐programming (DP) recurrence and implement it via an A*‐style search.  Define the cost‐to‐come
\begin{equation}
  g(v) \;=\;\min_{\pi: s\to v}\Bigl\{\sum_{(i\to j)\in \pi}\bigl(d(i,j) \;+\;\gamma\,F_{\mathrm{total}}(x_j,y_j)\bigr)\Bigr\},
  \label{costFunc}
\end{equation}
where \(g(s)=0\) and \(g(v)=+\infty\) for all \(v\ne s\) initially.  The one‐step DP update is
\begin{equation}
  g(v)
  \;\leftarrow\;
  \min_{(u\to v)\in E}\bigl\{\,g(u)+d(u,v)+\gamma\,F_{\mathrm{total}}(x_v,y_v)\bigr\}.
\end{equation}
To focus the DP on promising regions, we equip each node \(v\) with a (possibly inadmissible) heuristic estimate \(h(v)\) and maintain a priority queue OPEN of frontier nodes ordered by
\begin{equation}
  f(v) \;=\; g(v)\;+\;w_1\,h(v).
\end{equation}
Repeatedly extract the node \(v^*=\arg\min_{v\in\OPEN}f(v)\), relax all outgoing edges \((v^*\to w)\) via the above DP update (inserting/updating \(w\) in OPEN), and continue until the goal \(t\) is popped.  The result \(g(t)\) is then the minimum joint‐cost under the chosen heuristics and weights.

\medskip
\noindent\textbf{Multi-Heuristic A* specialization.}
To recover the two‐queue strategy of Multi-Heuristic A*, split the search into
\begin{itemize}
  \item An \emph{anchor} OPEN$_0$ using the admissible Euclidean heuristic
    \[
      h_0(v)=\lVert \mathbf{x}_v - \mathbf{x}_t\rVert_2,
    \]
    with key \(f_0(v)=g(v)+w_1\,h_0(v)\).
  \item An \emph{informed} OPEN$_1$ using the repulsive‐potential heuristic
    \[
      h_1(v)=F_{\mathrm{total}}(x_v,y_v),
    \]
    with key \(f_1(v)=g(v)+w_1\,h_1(v)\).
\end{itemize}
At each step, compare
\[
  \min_{v\in\OPEN_1} f_1(v)
  \;\le\;
  w_2\,\min_{v\in\OPEN_0} f_0(v),
\]
expanding from the corresponding queue.  This lazy DP update—guided by two heuristics—ensures both bounded‐suboptimality (via the anchor) and accelerated convergence (via the informed priority queue) \cite{MHA*}.
It is worth noting that if the potential field is zero throughout the map, then the multi-heuristic approach reduces to the classical A* algorithm, as shown in \cite{amani2025safe}. The pseudo-code of the algorithm can be seen in Algorithm \ref{alg:mh-astar}.
\begin{algorithm}[h]
\caption{Multi–Heuristic A* with Repulsive Potential} \label{alg:mh-astar}
\scriptsize
\begin{algorithmic}[1]
  \Require Grid $G$, start $s$, goal $t$, weights $w_1>0$, $w_2\ge1$, $\gamma\ge0$
  \State Init OPEN$_0$, OPEN$_1$, CLOSED$_{anc}$, CLOSED$_{inad}$
  \For{$n\in G$} $g(n)\gets\infty$, $p(n)\gets$ null \EndFor
  \State $g(s)\gets0$; enqueue $s$ in both OPENs
  \While{OPEN$_0\neq\emptyset$}
    \State $u_0\gets$ top(OPEN$_0$); $u_1\gets$ top(OPEN$_1$)
    \If{Key$(u_1,1)\le w_2$Key$(u_0,0)$} $u\gets u_1$, add to CLOSED$_{inad}$
    \Else $u\gets u_0$, add to CLOSED$_{anc}$ \EndIf
    \If{$g(t)\le$Key$(u,\cdot)$} \textbf{break} \EndIf
    \For{neighbor $v$ of $u$ if walkable}
      \State $\hat g=g(u)+\|u-v\|_2+\gamma v.p$
      \If{$\hat g<g(v)$} $g(v)\gets\hat g$, $p(v)\gets u$, enqueue in OPENs \EndIf
    \EndFor
  \EndWhile
  \State \Return path from $s$ to $t$
\end{algorithmic}
\end{algorithm}

\subsection{Beta–Bernoulli Fusion of LLM Danger Readings}

We treat LLMs as semantic sensors that assess obstacle danger. For each obstacle, we model the fundamental question "Is this obstacle dangerous?" as a binary classification problem. This binary danger state follows a Bernoulli distribution with unknown probability $p_i$, for which we maintain a Beta conjugate prior, which is an approach analogous to Bayesian occupancy grid mapping \cite{Thrun2002ProbabilisticRobotics}.  
\par

The Beta-Bernoulli framework serves two critical purposes. First, the Beta distribution is the conjugate prior for Bernoulli observations, enabling closed-form Bayesian updates. Second, and crucially for LLM integration, the Beta posterior never reaches exactly 0 or 1 when both parameters remain positive, preventing the numerical instabilities that would arise from LLM hallucinations producing extreme confidence values.
\par

Rather than directly using the LLM's danger score \(\hat p_i\in[0,1]\) as a probability (which could cause posterior collapse if \(\hat p_i\) = 0 or 1), we interpret it as evidence accumulated from NN
N virtual Bernoulli trials. This pseudo-count interpretation ensures that even extreme LLM outputs result in finite Beta parameter updates, maintaining numerical stability throughout sequential prompt processing.

Initially, each obstacle’s danger probability has a uniform Beta prior:
\[
p_i \sim \mathrm{Beta}(\alpha_i,\beta_i),
\quad
\alpha_i=\beta_i=1.
\]
The Beta density is
\begin{align}
f(x;\alpha,\beta)
&= \frac{x^{\alpha-1}(1-x)^{\beta-1}}{\mathrm{Beta}(\alpha,\beta)},
\quad 0 < x < 1,\\
\mathrm{Beta}(\alpha,\beta)
&= \frac{\Gamma(\alpha)\,\Gamma(\beta)}{\Gamma(\alpha+\beta)}.
\end{align}

Upon receiving prompt \(P\), we ask the LLM for each obstacle’s “danger score” \(\hat p_i\in[0,1]\).  Interpreting \(\hat p_i\) as the mean of \(N\) Bernoulli trials, we form pseudo-counts
\[
a_i = N\,\hat p_i,
\quad
b_i = N\,(1-\hat p_i),
\]
and update
\[
\alpha_i \leftarrow \alpha_i + a_i,
\quad
\beta_i  \leftarrow \beta_i  + b_i.
\]
The posterior remains
\[
p_i \mid \hat p_i \sim \mathrm{Beta}(\alpha_i,\beta_i),
\]
with expectation
\[
\mathbb{E}[\,p_i\mid \hat p_i\,]
= \frac{\alpha_i}{\alpha_i+\beta_i}.
\]
Finally, we scale each obstacle’s base repulsive gain \(g_i^0\) by this mean:
\begin{equation}
\lambda_i \;\leftarrow\; 
\frac{\alpha_i}{\alpha_i + \beta_i} \, \lambda_{i}^0.
\label{eq:gain_update}
\end{equation}
Note that as long as \(\alpha >0 \) and \(\beta>0\), then the posterior mean is guaranteed to be between 0 and 1. This prevents the expectation from "locking in" on probabilities 0 and 1, providing guaranteed numerical stability in chained prompt settings.
This Beta–Bernoulli scheme fuses designer priors with LLM observations in closed form, offers a clear “trust” knob \(N\), and supports chaining multiple prompts without numerical issues. Note that N here plays the role of a “virtual sample size,” controlling how much weight each LLM‐derived score carries when forming the likelihood.
\par

\begin{figure}[htbp]
  \centering
  \begin{subfigure}[b]{0.48\columnwidth}
    \includegraphics[width=\linewidth]{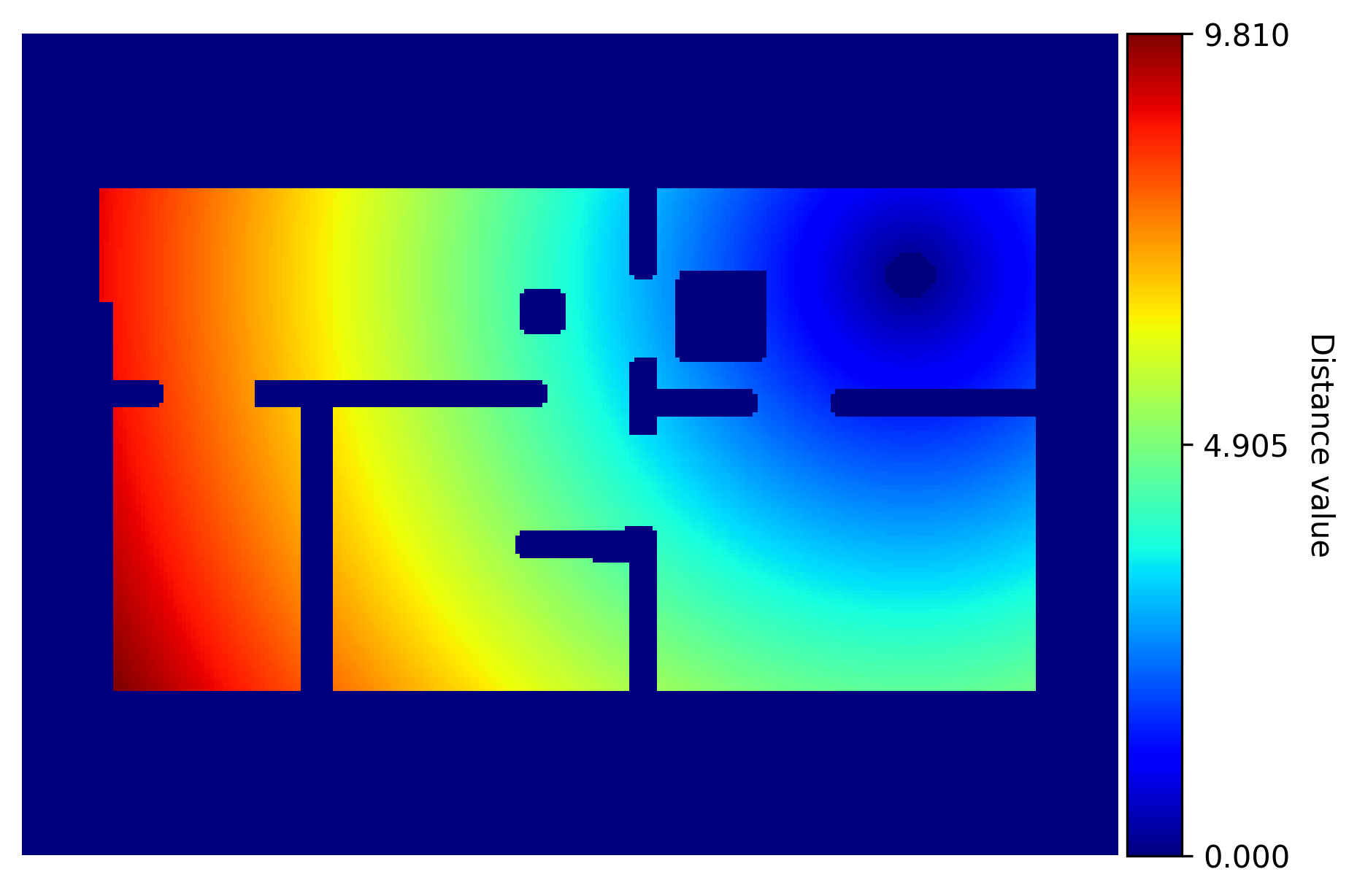}
    \caption{Distance cost}
    \label{fig:avoidWSDist}
  \end{subfigure}
  \hfill
  \begin{subfigure}[b]{0.48\columnwidth}
    \includegraphics[width=\linewidth]{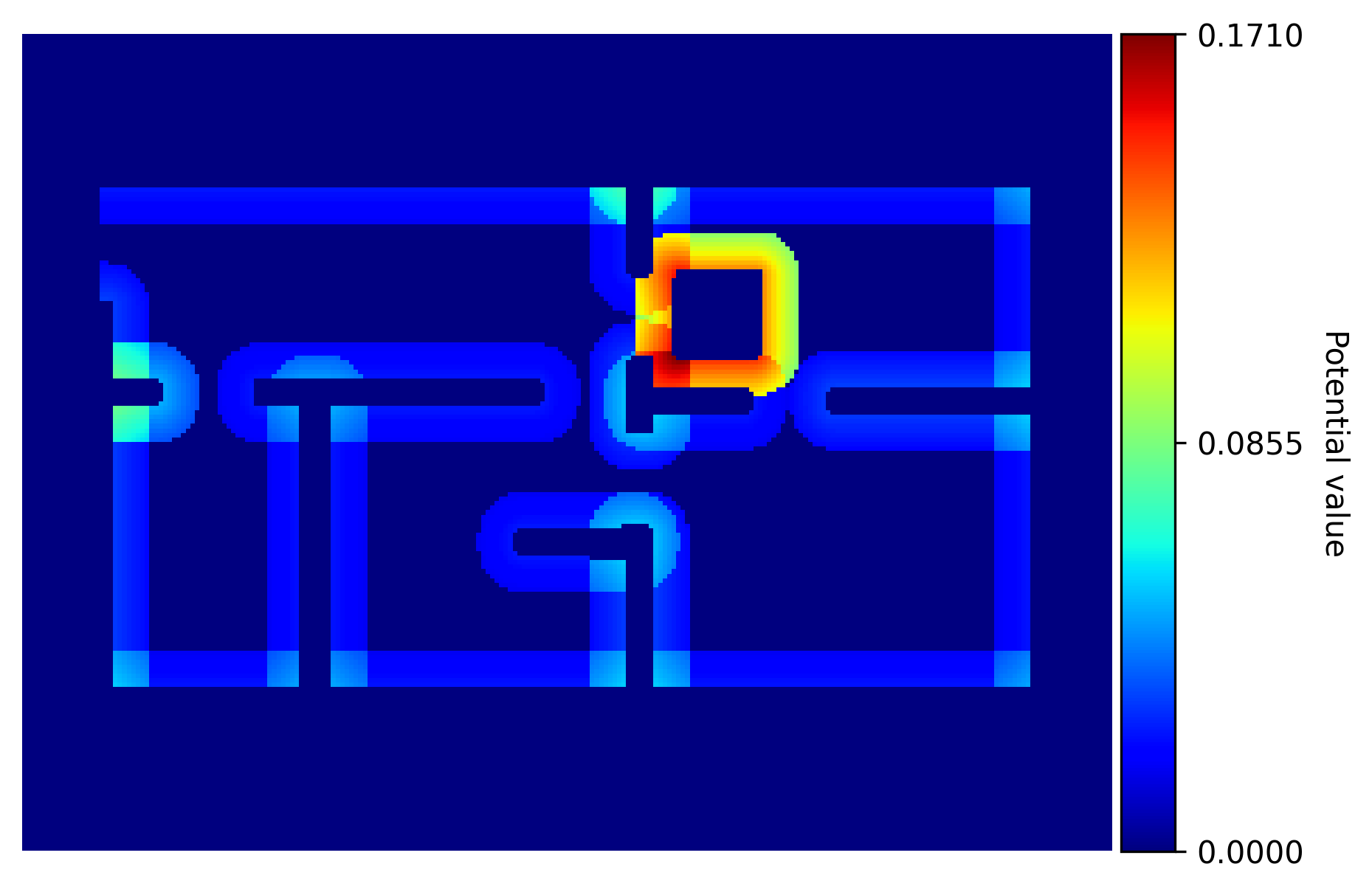}
    \caption{Repulsive potential cost}
    \label{fig:avoidWSRep}
  \end{subfigure}

  \medskip 

  \begin{subfigure}[b]{0.48\columnwidth}
    \includegraphics[width=\linewidth]{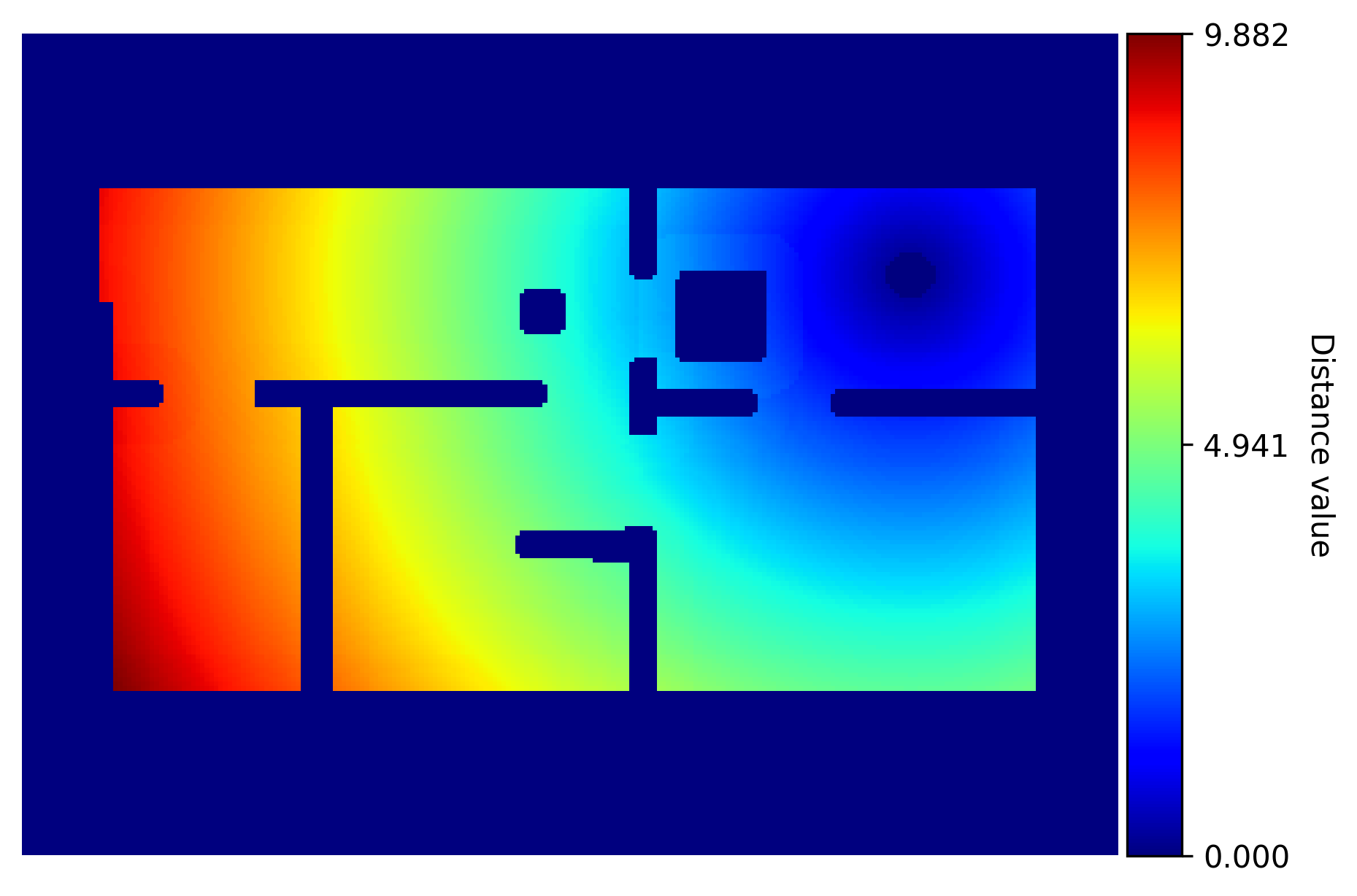}
    \caption{Combined cost}
    \label{fig:avoidWSCost}
  \end{subfigure}
  \hfill
  \begin{subfigure}[b]{0.48\columnwidth}
    \includegraphics[width=\linewidth]{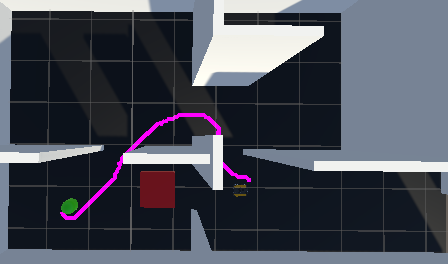}
    \caption{Optimal path}
    \label{fig:avoidWSResult}
  \end{subfigure}

  \caption{Cost‐map components associated with the mission planning (cf.\ Equation~\ref{costFunc}): 
  (a) distance‐to‐goal cost, 
  (b) obstacle repulsive cost, 
  (c) the combined cost used for planning, and 
  (d) the resulting optimal path given the minimized cost.}
  \label{fig:avoidWSCostmaps}
\end{figure}

Given the cost function defined in Equation \ref{costFunc}, we can visualize the individual components and the combined cost map, as can be seen in Figure \ref{fig:avoidWSCostmaps}. Figure \ref{fig:avoidWSRep} represents the Euclidean distance towards the goal. Figure \ref{fig:avoidWSRep} represents the object avoidance cost, Figure \ref{fig:avoidWSCost} the combination of the two costs, which results in the path optimization realized in Figure \ref{fig:avoidWSResult}.

\section{Experiments}

\begin{figure}[htbp]
  \centering
  \includegraphics[width=1\columnwidth]{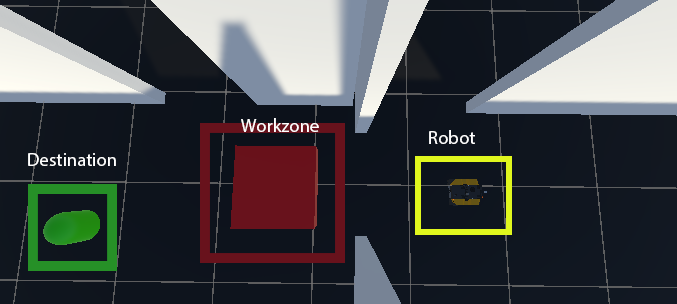}
  \caption{Work area represented by a red cube in Unity.}
  \label{fig:red-cube}
\end{figure}

We experiment with NL prompts in a setting where a work area has been determined in the digital twin. The BIM is sourced from the floor plan of the SDSU DiCE Lab architectural Revit file that has been transformed into an importable FBX file. In the Unity environment, the work area of interest is visualized as a red cube, the robot is modeled as a differential-drive unmanned ground vehicle (UGV), and the destination is shown as a green capsule. These objects are shown in Figure \ref{fig:red-cube}
\par
We used GPT-3.5-Turbo to estimate a danger coefficient for every obstacle. The model was invoked through a single, structured prompt (see Supplementary Listing 1 in our GitHub repository) that describes the obstacle’s salient features and asks the model to return a normalized scalar indicating relative risk. All API parameters were left at their default values: \texttt{temperature = 1.0}, $top_p = 1.0$, \texttt{frequency penalty = 0}, and \texttt{presence penalty = 0}.

\subsection{Semantic understanding of avoidance given probability of collisions with humans}
\begin{figure}[htbp]
  \centering
  \begin{subfigure}[b]{0.5\columnwidth}
    \includegraphics[width=\linewidth]{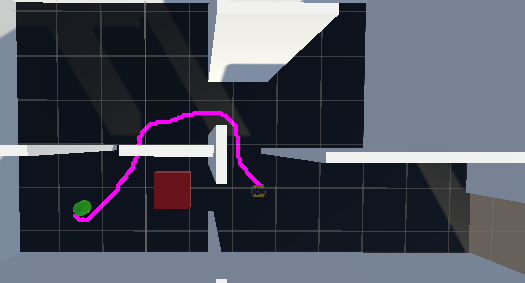}
    \caption{Busy prompt}
    \label{fig:busy}
  \end{subfigure}
  \hfill
  \begin{subfigure}[b]{0.5\columnwidth}
    \includegraphics[width=\linewidth]{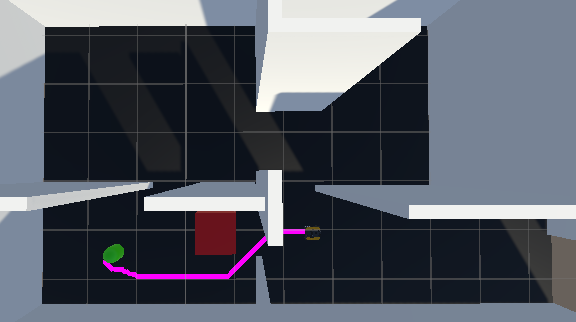}
    \caption{Empty prompt}
    \label{fig:empty}
  \end{subfigure}
  \hfill
  \begin{subfigure}[b]{0.5\columnwidth}
    \includegraphics[width=\linewidth]{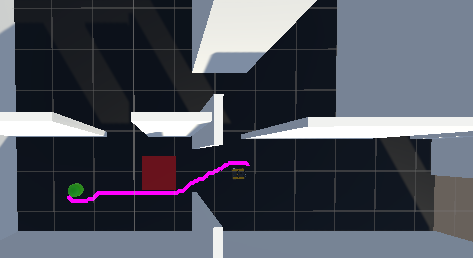}
    \caption{A* baseline}
    \label{fig:astar}
  \end{subfigure}

  \caption{Comparison of planner behavior under (a) busy, (b) empty, and (c) A* baseline prompts.}
  \label{fig:workzone_paths}
\end{figure}

Here, we show two separate prompts that are provided to the robot regarding the environment.
\begin{itemize}
  \item \enquote{The work zone is empty today; proceed to your destination.}
  \item \enquote{The work zone is busy today; proceed to your destination.}
\end{itemize}
Next, we evaluate how different the planning algorithm performs with respect to the provided NL command. The results are shown in Figure \ref{fig:workzone_paths}, where substantial qualitative differences between the chosen paths can be observed.


\begin{table}[htbp]
  \centering
  \caption{Comparison of Path Metrics and Bernoulli Updates}
  \label{tab:workzone}
  \begin{tabular}{
    l
    S[table-format=2.2]  
    S[table-format=2.2]  
    S[table-format=2.2]  
  }
  \toprule
   {Prompt} & {Empty} & {Busy} & {A* Baseline} \\
  \midrule
  \multicolumn{4}{c}{\textbf{Path Metrics}} \\
  \midrule
  Length (units)            & 5.07 & 7.00 & 4.910 \\
  Min.\ obstacle dist.\ (m) & 0.137 & 0.481 & 0.100 \\
  Avg.\ obstacle dist.\ (m) & 0.537 & 0.631 & 0.490 \\
  \midrule
  \multicolumn{4}{c}{\textbf{Beta–Bernoulli Updates}} \\
  \midrule
  Workstations (post.)      & 0.21 & 0.86 & {—} \\
  Wall (post.)              & 0.21 & 0.29 & {—} \\
  \bottomrule
  \end{tabular}
\end{table}

\subsection{Semantic experiments with regard to collisions with materials or hazardous objects}
In this experiment, we use the exact same positions and setup as in the previous one, but we analyze the industrial semantic understanding within the model. We focus on the understanding of Mechanical, Electrical, and Plumbing (MEP) semantics within the construction schedule. It is expected to receive either the exact or similar path behaviors as the person's collision semantics. Specifically, we provide the following prompts to the UGV:.
\begin{itemize}
  \item \enquote{Workzone has completed electrical conduits installation according to the schedule, go to the destination.}
  \item \enquote{Workzone is undergoing electrical conduits installation according to the schedule, go to the destination.}
\end{itemize}

\begin{figure}[!t]
  \centering
  \begin{subfigure}[b]{1\columnwidth}
    \includegraphics[width=\columnwidth]{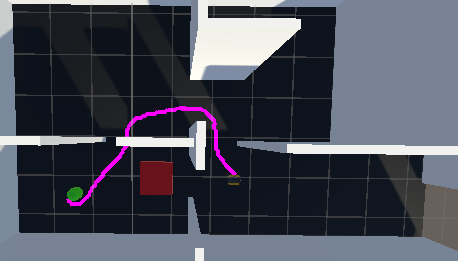}
    \caption{Path chosen for the prompt signifying ongoing MEP installation.}
    \label{fig:busy_MEP}
  \end{subfigure}
  \hfill
  \begin{subfigure}[b]{1\columnwidth}
    \includegraphics[width=\columnwidth]{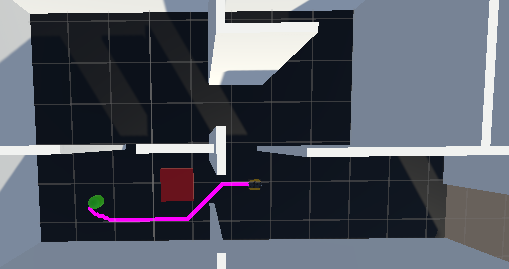}
    \caption{Path chosen for the prompt signifying no ongoing MEP installation.}
    \label{fig:empty_MEP}
  \end{subfigure}
  \caption{Comparison of planner behavior under (a) active and (b) inactive work-zone commands.}
  \label{fig:workzone_paths_MEP}
\end{figure}

Resulting paths are shown Figure \ref{fig:workzone_paths_MEP}.


\begin{table}[htbp]
  \centering
  \caption{Comparison of Path Metrics and Bernoulli Updates for MEP and A* Baseline}
  \label{tab:workzone_MEP}
  \sisetup{
    table-format=1.3,     
    detect-weight=true,    
    table-number-alignment = center
  }
  \begin{tabular}{
    l
    S  
    S  
    S  
  }
    \toprule
      & {MEP installed} & {Ongoing MEP} & {A* Baseline} \\
    \midrule
    \multicolumn{4}{c}{\textbf{Path Metrics}} \\
    \midrule
    Length (units)            & 4.750 & 7.000 & {4.910} \\
    Min.\ obstacle dist.\,(m) & 0.137 & 0.481 & {0.1} \\
    Avg.\ obstacle dist.\,(m) & 0.554 & 0.629 & {0.490} \\
    \midrule
    \multicolumn{4}{c}{\textbf{Beta–Bernoulli Updates}} \\
    \midrule
    Workstations (post.)      & 0.36  & 0.71  & {—} \\
    Wall (post.)              & 0.21  & 0.57  & {—} \\
    \bottomrule
  \end{tabular}
\end{table}

\subsection{Object specific avoidance}

Here, we approach obstacle-specific avoidance from understanding context. We experiment with the understanding of the model of consequences of the danger implicitly, without just thinking about collisions. A three-object environment is designed where the top is a storage unit, the middle is a recently cemented floor area, and the bottom is a welding station. We experiment with prompts implying certain characteristics unique to the object. We present the robot with two separate prompts:
\begin{enumerate}
    \item "Workstation is busy today, go to your destination" 
    \item "Workstation is empty today, go to your destination" 
\end{enumerate}
Note that the only difference between the prompts are the words "busy" and "empty".
\subsection{Semantic understanding of avoidance given probability of collisions with humans}
In this experiment, we provide two separate prompts to the robot regarding the environment.
\begin{itemize}
  \item \enquote{The environment is pretty empty and all of the work is completed, go to the destination}
  \item \enquote{The works zone is empty, but we poured cement 40 minutes ago, go to the destination}
\end{itemize}

\begin{figure}[htbp]
  \centering
  \begin{subfigure}[b]{0.5\columnwidth}
    \includegraphics[width=\linewidth]{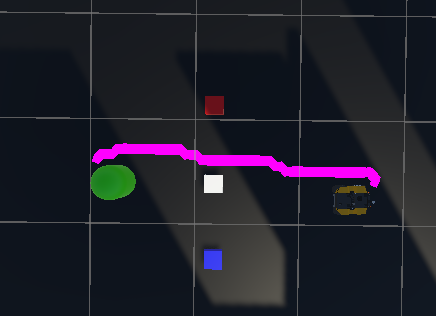}
    \caption{Safe path}
    \label{fig:busy_CEMENT}
  \end{subfigure}
  \hfill
  \begin{subfigure}[b]{0.5\columnwidth}
    \includegraphics[width=\linewidth]{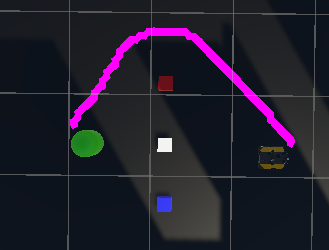}
    \caption{Freshly poured cement}
    \label{fig:empty_CEMENT}
  \end{subfigure}
  \hfill
  \begin{subfigure}[b]{0.5\columnwidth}
    \includegraphics[width=\linewidth]{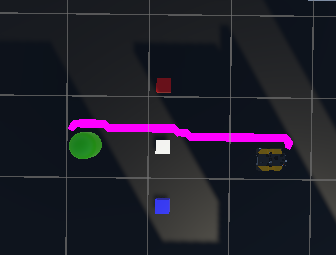}
    \caption{A* baseline}
    \label{fig:ASTAR_CEMENT}
  \end{subfigure}

  \caption{Comparison of planner behavior under (a) safe, (b) freshly poured cement, and (c) A* baseline. The white box represents the cement area, the red box the storage unit, and the blue box a welding station.}
  \label{fig:cement_paths}
\end{figure}

Here, the algorithm is expected to understand that freshly poured cement is something to avoid, given the dangers associated with traversing over it. Resulting paths are shown in Figure \ref{fig:cement_paths}.
\subsection{Ablation Study}
LLMs are susceptible to hallucinating. In order to examine the effects of the LLM's hallucinations and the posteriors associated with a selection of BIM families, we ran 100 tests and calculated posterior means and standard deviations. The LLM was prompted to return danger likelihoods given only the object name. We can see reasonable posteriors and relatively stable standard deviations in Table \ref{tab:beta_posteriors_full}. This shows that the GPT 3.5 Turbo is a suitable model for this semantic analysis of this task. If the probabilistic formulation is different and the LLM is used in another setting, the LLM needs to be evaluated rigorously to meet the requirements of the task. 
\begin{table}[ht]
  \centering
  \caption{Beta–Bernoulli Posterior Statistics over 100 Runs}
  \label{tab:beta_posteriors_full}
  \begin{tabular}{@{}lc@{}}
    \toprule
    \textbf{Object}       & \textbf{Posterior Mean $\pm$ Standard Deviation} \\
    \midrule
    Walls                 & 0.4121 $\pm$ 0.1160 \\
    Grinder               & 0.6929 $\pm$ 0.0627 \\
    Chainsaw              & 0.7471 $\pm$ 0.0434 \\
    Chair                 & 0.2186 $\pm$ 0.0170 \\
    Floor‐Cement          & 0.4186 $\pm$ 0.0639 \\
    Floor‐Cement-Dry          & 0.3479 $\pm$ 0.0697 \\
    Welding‐Station       & 0.6143 $\pm$ 0.0495 \\
    Storage               & 0.4121 $\pm$ 0.0775 \\
    \bottomrule
  \end{tabular}
\end{table}

\begin{figure}[htbp]
  \centering
  \includegraphics[width=\columnwidth]{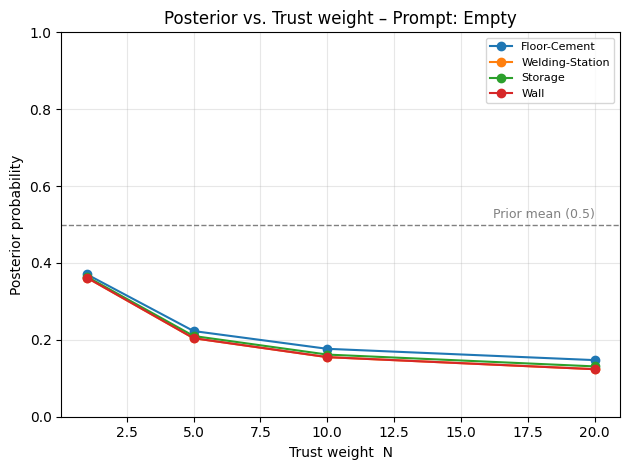}
  \caption{Ablation statistics from varying the trust knob given an empty prompt (50 queries).}
  \label{fig:emptyStats}
\end{figure}

\begin{figure}[htbp]
  \centering
  \includegraphics[width=\columnwidth]{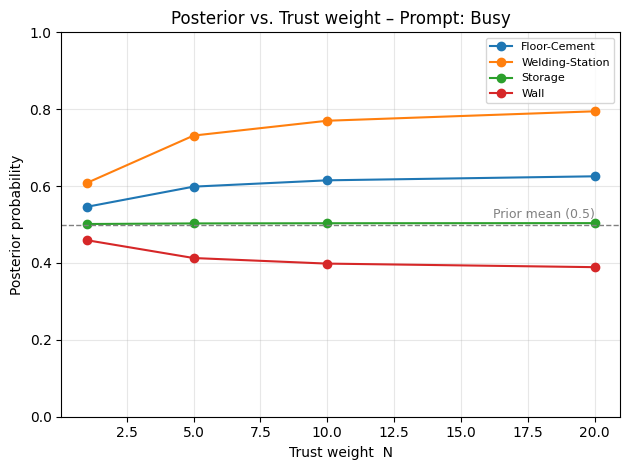}
  \caption{Ablation statistics from varying the trust knob given a busy prompt (50 queries).}
  \label{fig:busyStats}
\end{figure}

We further investigated the ablation of the trust knob on the effect of the posterior after the Bayesian update. The results can be seen in Figures \ref{fig:emptyStats} and \ref{fig:busyStats}. We can see that as the trust knob increases, we see the likelihood affect the posterior probability more drastically. We can see in the busy scenario, the LLM weighs the dangers much differently regarding activity and potential dangers. Since the LLM does not know if the cement is dry, it increases its danger belief; however, not to an incredibly high degree. We can see the welding station is avoided at a much more aggressive rate as opposed to anything else. This is not the same when we encounter an empty worksite since the LLM converges to a dangerous belief that can only be due to current map offsets and not real-time mapping errors under dynamic environments.

\begin{table}
  \centering
  \caption{Comparison of Path Metrics and Bernoulli Updates for Cement Prompts and A* Baseline}
  \label{tab:workzone_safecement}
  \sisetup{
    table-format=1.3,             
    detect-weight=true,
    table-number-alignment = center
  }
  \begin{tabular}{l S S S}
    \toprule
      & {Dried Cement} & {Wet Cement} & {A* Baseline} \\
    \midrule
    \multicolumn{4}{c}{\textbf{Path Metrics}} \\
    \midrule
    Path length (units)            & 2.840 & 3.950 & {2.800} \\
    Min.\ obstacle dist.\,(m)      & 0.138 & 0.438 & {0.138} \\
    Avg.\ obstacle dist.\,(m)      & 0.579 & 0.684 & {0.593} \\
    \midrule
    \multicolumn{4}{c}{\textbf{Beta–Bernoulli Updates}} \\
    \midrule
    Floor–Cement (post.)           & 0.21  & 0.71  & {—} \\
    Welding Station (post.)        & 0.71  & 0.29  & {—} \\
    Storage (post.)                & 0.21  & 0.21  & {—} \\
    \bottomrule
  \end{tabular}
\end{table}
\section{Discussion}
Our approach yields a formally verifiable method for incorporating semantic and safety knowledge into robotic path-planning. Unlike VLA models, which offer impressive empirical performance but limited formal guarantees, our framework retains the provable optimality bounds typical of classical algorithms while still leveraging LLMs for richer scene understanding. These guarantees make the method attractive for mission-critical, high-stakes deployments in which purely probabilistic planners may fail to meet required safety or performance constraints.
\par

There are interesting observations that we see in regards to how prompts and the LLM interact. Since the LLM is instructed that it is to assign dangers given the danger of robot navigation collisions, the individual scalings are affected overall by the prompt. We can see in the presence of an empty worksite, all danger readings decrease \ref{fig:emptyStats}. However, in the presence of a dynamic and busy job site, the scaling is more object-specific. For example, we would see a welding station have a bigger increase in danger given the prompt as opposed to a storage unit, since the LLM detects that a welding station will be more dynamic under busy conditions than the storage unit. We can also see that the cemented floor is also increasing in danger readings \ref{fig:busyStats}. This is a promising result given the LLM's ability to understand underlying semantics because the BIM model does not specify if the cement has been dried and whether or not it is safe to drive over. 

\par
The proposed methodology provides a lightweight, transparent way to fuse NLP‐derived sentiment into motion planning. So far, we have only plugged it into a simple cost‐minimization scheme via a single repulsive potential field. But this rigid coupling can fail in degenerate cases—for example, when there is only a single collision‐free corridor from start to goal, no amount of repulsive “push” will change the chosen path, and the policy reduces to mere heuristic tweaks.

To overcome this limitation, we must explore richer planning paradigms. One option is multi‐objective optimization, blending repulsion, path length, and smoothness into a Pareto framework. Another is hierarchical planning, where Bayesian‐updated danger scores inform subgoal selection rather than raw costs. Sampling‐based planners (e.g., RRT*, PRM) can also be biased toward high‐confidence regions according to the posterior mean. Integrating these strategies will let us leverage LLM “trust” scores in a more granular, context‐sensitive way.

\par
This approach is of high importance for integrating construction robots into real-world applications. By providing guaranteeable and closed-form methods, we can ensure predictable behaviors in both the robot’s actions and overall mission planning. Such guaranteeable approaches are in high demand in mission-critical and safety-critical settings.

Even in the worst case of hallucinations, the model falls back to the classical A* algorithm, which is already widely used in industry. Thus, the worst-case performance of this method is equivalent to A*, ensuring reliability. While this paper uses A*, its main contribution is not presenting an A*-like algorithm. Instead, the core contribution lies in creating a costmap formulation that leverages BIM semantics, human prompt sentiment analysis, and repulsive potential concepts. This enables probabilistic and incremental costmap generation as the system learns more about its environment through prompts. A* is applied here solely to solve the cost minimization problem. Other label correcting algorithms could be substituted and may perform differently or even better, depending on their implementation and setting.
\section{Conclusion and Future Work}
In this paper, we approached the integration of both explicit and implicit NL path modification in danger-aware path planning within BIM environments. In the future, we plan to consider different use cases and cost functions to minimize for different mission plans outside the scope of collision-free path planning. We also plan to use similar concepts, such as the Bayesian interpretation of the LLM with prompts, to encapsulate spatial understanding for robots as well.
\par
We plan to extend this approach to 3D path planning and mapping refinement, investigating optimal strategies that reduce computational load and enhance material understanding in three dimensions. Additionally, recent advances demonstrate that integrating language semantics can significantly improve robotic mapping performance. We intend to explore these promising directions to further leverage LLM-driven insights in complex spatial domains.

\bibliography{ascexmpl-new}
\vspace{12pt}

\end{document}